# GENERATIVE MULTI-ADVERSARIAL NETWORKS

**Ishan Durugkar**[*]**, Ian Gemp**[*]**, Sridhar Mahadevan**
College of Information and Computer Sciences
University of Massachusetts, Amherst
Amherst, MA 01060, USA
`{idurugkar,imgemp,mahadeva}@cs.umass.edu`

## ABSTRACT

Generative adversarial networks (GANs) are a framework for producing a generative model by way of a two-player minimax game. In this paper, we propose the *Generative Multi-Adversarial Network* (GMAN), a framework that extends GANs to multiple discriminators. In previous work, the successful training of GANs requires modifying the minimax objective to accelerate training early on. In contrast, GMAN can be reliably trained with the original, untampered objective. We explore a number of design perspectives with the discriminator role ranging from formidable adversary to forgiving teacher. Image generation tasks comparing the proposed framework to standard GANs demonstrate GMAN produces higher quality samples in a fraction of the iterations when measured by a pairwise GAM-type metric.

## 1 INTRODUCTION

Generative adversarial networks (Goodfellow et al. (2014)) (GANs) are a framework for producing a generative model by way of a two-player minimax game. One player, the generator, attempts to generate realistic data samples by transforming noisy samples, $z$, drawn from a simple distribution (e.g., $z \sim \mathcal{N}(0, 1)$) using a transformation function $G_\theta(z)$ with learned weights, $\theta$. The generator receives feedback as to how realistic its synthetic sample is from another player, the discriminator, which attempts to discern between synthetic data samples produced by the generator and samples drawn from an actual dataset using a function $D_\omega(x)$ with learned weights, $\omega$.

The GAN framework is one of the more recent successes in a line of research on adversarial training in machine learning (Schmidhuber (1992); Bagnell (2005); Ajakan et al. (2014)) where games between learners are carefully crafted so that Nash equilibria coincide with some set of desired optimality criteria. Preliminary work on GANs focused on generating images (e.g., MNIST (LeCun et al. (1998)), CIFAR (Krizhevsky (2009))), however, GANs have proven useful in a variety of application domains including learning censored representations (Edwards & Storkey (2015)), imitating expert policies (Ho & Ermon (2016)), and domain transfer (Yoo et al. (2016)). Work extending GANs to semi-supervised learning (Chen et al. (2016); Mirza & Osindero (2014); Gauthier (2014); Springenberg (2015)), inference (Makhzani et al. (2015); Dumoulin et al. (2016)), feature learning (Donahue et al. (2016)), and improved image generation (Im et al. (2016); Denton et al. (2015); Radford et al. (2015)) have shown promise as well.

Despite these successes, GANs are reputably difficult to train. While research is still underway to improve training techniques and heuristics (Salimans et al. (2016)), most approaches have focused on understanding and generalizing GANs theoretically with the aim of exploring more tractable formulations (Zhao et al. (2016); Li et al. (2015); Uehara et al. (2016); Nowozin et al. (2016)).

In this paper, we theoretically and empirically justify generalizing the GAN framework to multiple discriminators. We review GANs and summarize our extension in Section 2. In Sections 3 and 4, we present our $N$-discriminator extension to the GAN framework (*Generative Multi-Adversarial Networks*) with several variants which range the role of the discriminator from formidable adversary to forgiving teacher. Section 4.2 explains how this extension makes training with the untampered minimax objective tractable. In Section 5, we define an intuitive metric (GMAM) to quantify GMAN performance and evaluate our framework on a variety of image generation tasks. Section 6 concludes with a summary of our contributions and directions for future research.

---

[*]Equal contribution

**Contributions**—To summarize, our main contributions are: **i)** a multi-discriminator GAN framework, GMAN, that allows training with the original, untampered minimax objective; **ii)** a generative multi-adversarial metric (GMAM) to perform pairwise evaluation of separately trained frameworks; **iii)** a particular instance of GMAN, GMAN*, that allows the generator to automatically regulate training and reach higher performance (as measured by GMAM) in a fraction of the training time required for the standard GAN model.

## 2 GENERATIVE ADVERSARIAL NETWORKS TO GMAN

The original formulation of a GAN is a minimax game between a generator, $G_\theta(z) : z \to x$, and a discriminator, $D_\omega(x) : x \to [0, 1]$,

$$\min_G \max_{D \in \mathcal{D}} V(D, G) = \mathbb{E}_{x \sim p_{data}(x)} \Big[ \log(D(x)) \Big] + \mathbb{E}_{z \sim p_z(z)} \Big[ \log(1 - D(G(z))) \Big], \quad (1)$$

where $p_{data}(x)$ is the true data distribution and $p_z(z)$ is a simple (usually fixed) distribution that is easy to draw samples from (e.g., $\mathcal{N}(0, 1)$). We differentiate between the function space of discriminators, $\mathcal{D}$, and elements of this space, $D$. Let $p_G(x)$ be the distribution induced by the generator, $G_\theta(z)$. We assume $D, G$ to be deep neural networks as is typically the case.

In their original work, Goodfellow et al. (2014) proved that given sufficient network capacities and an oracle providing the optimal discriminator, $D^* = arg \max_D V(D, G)$, gradient descent on $p_G(x)$ will recover the desired globally optimal solution, $p_G(x) = p_{data}(x)$, so that the generator distribution exactly matches the data distribution. In practice, they replaced the second term, $\log(1 - D(G(z)))$, with $-\log(D(G(z)))$ to enhance gradient signals at the start of the game; note this is no longer a zero-sum game. Part of their convergence and optimality proof involves using the oracle, $D^*$, to reduce the minimax game to a minimization over $G$ only:

$$\min_G V(D^*, G) = \min_G \Big\{ C(G) = -\log(4) + 2 \cdot JSD(p_{data}||p_G) \Big\} \quad (2)$$

where $JSD$ denotes Jensen-Shannon divergence. Minimizing $C(G)$ necessarily minimizes $JSD$, however, we rarely know $D^*$ and so we instead minimize $V(D, G)$, which is only a lower bound.

This perspective of minimizing the distance between the distributions, $p_{data}$ and $p_G$, motivated Li et al. (2015) to develop a generative model that matches all moments of $p_G(x)$ with $p_{data}(x)$ (at optimality) by minimizing maximum mean discrepancy (MMD). Another approach, EBGAN, (Zhao et al. (2016)) explores a larger class of games (non-zero-sum games) which generalize the generator and discriminator objectives to take real-valued "energies" as input instead of probabilities. Nowozin et al. (2016) and then Uehara et al. (2016) extended the $JSD$ perspective on GANs to more general divergences, specifically $f$-divergences and then Bregman-divergences respectively.

In general, these approaches focus on exploring fundamental reformulations of $V(D, G)$. Similarly, our work focuses on a fundamental reformulation, however, our aim is to provide a framework that accelerates training of the generator to a more robust state irrespective of the choice of $V$.

### 2.1 GMAN: A MULTI-ADVERSARIAL EXTENSION

We propose introducing multiple discriminators, which brings with it a number of design possibilities. We explore approaches ranging between two extremes: 1) a more discriminating $D$ (better approximating $\max_D V(D, G)$) and 2) a $D$ better matched to the generator's capabilities. Mathematically, we reformulate $G$'s objective as $\min_G \max F(V(D_1, G), \ldots, V(D_N, G))$ for different choices of $F$ (see Figure 1). Each $D_i$ is still expected to independently maximize its own $V(D_i, G)$ (i.e. no cooperation). We sometimes abbreviate $V(D_i, G)$ with $V_i$ and $F(V_1, \ldots, V_N)$ with $F_G(V_i)$.

## 3 A FORMIDABLE ADVERSARY

Here, we consider multi-discriminator variants that attempt to better approximate $\max_D V(D, G)$, providing a harsher critic to the generator.

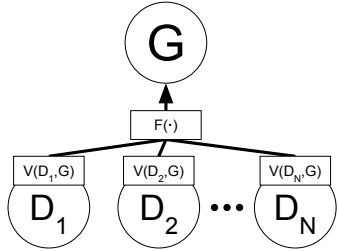

Figure 1: (GMAN) The generator trains using feedback aggregated over multiple discriminators. If $F := \max$, $G$ trains against the best discriminator. If $F := \texttt{mean}$, $G$ trains against an ensemble. We explore other alternatives to $F$ in Sections 4.1 & 4.4 that improve on both these options.

### 3.1 MAXIMIZING V(D,G)

For a fixed $G$, maximizing $F_G(V_i)$ with $F = \max$ and $N$ randomly instantiated copies of our discriminator is functionally equivalent to optimizing $V$ (e.g., stochastic gradient ascent) with random restarts in parallel and then presenting $\max_{i \in \{1,...,N\}} V(D_i, G)$ as the loss to the generator —a very pragmatic approach to the difficulties presented by the non-convexity of $V$ caused by the deep net. Requiring the generator to minimize the $\max$ forces $G$ to generate high fidelity samples that must hold up under the scrutiny of all $N$ discriminators, each potentially representing a distinct max.

In practice, $\max_{D_i \in \mathcal{D}} V(D_i, G)$ is not performed to convergence (or global optimality), so the above problem is oversimplified. Furthermore, introducing $N$ discriminators affects the dynamics of the game which affects the trajectories of the discriminators. This prevents us from claiming $\max\{V_1(t),\ldots,V_N(t)\} > \max\{V_1'(t)\} \; \forall t$ even if we initalize $D_1(0) = D_1'(0)$ as it is unlikely that $D_1(t) = D_1'(t)$ at some time $t$ after the start of the game.

### 3.2 BOOSTING

We can also consider taking the $\max$ over $N$ discriminators as a form of boosting for the discriminator's online classification problem (online because $G$ can produce an infinite data stream). The *boosted* discriminator is given a sample $x_t$ and must predict whether it came from the generator or the dataset. The booster then makes its prediction using the predictions of the $N$ weaker $D_i$.

There are a few differences between taking the $\max$ (case 1) and online boosting (case 2). In case 1, our booster is limited to selecting a single weak discriminator (i.e. a pure strategy), while in case 2, many boosting algorithms more generally use linear combinations of the discriminators. Moreover, in case 2, a booster must make a prediction before receiving a loss function. In case 1, we assume access to the loss function at prediction time, which allows us to compute the $\max$.

It is possible to train the weak discriminators using boosting and then ignore the booster's prediction by instead presenting $\max\{V_i\}$. We explore both variants in our experiments, using the adaptive algorithm proposed in Beygelzimer et al. (2015). Unfortunately, boosting failed to produce promising results on the image generation tasks. It is possible that boosting produces too strong an adversary for learning which motivates the next section. Boosting results appear in Appendix A.7.

## 4 A FORGIVING TEACHER

The previous perspectives focus on improving the discriminator with the goal of presenting a better approximation of $\max_{\mathcal{D}} V(D, G)$ to the generator. Our next perspective asks the question, "Is $\max_{\mathcal{D}} V(D, G)$ too harsh a critic?"

### 4.1 *Soft*-DISCRIMINATOR

In practice, training against a far superior discriminator can impede the generator's learning. This is because the generator is unlikely to generate any samples considered "realistic" by the discriminator's standards, and so the generator will receive uniformly negative feedback. This is problem-atic because the information contained in the gradient derived from negative feedback only dictates where to drive down $p_G(x)$, not specifically where to increase $p_G(x)$. Furthermore, driving down $p_G(x)$ necessarily increases $p_G(x)$ in other regions of $\mathcal{X}$ (to maintain $\int_{\mathcal{X}} p_G(x) = 1$) which may or may not contain samples from the true dataset (*whack-a-mole* dilemma). In contrast, a generator is more likely to see positive feedback against a more lenient discriminator, which may better guide a generator towards amassing $p_G(x)$ in approximately correct regions of $\mathcal{X}$.

For this reason, we explore a variety of functions that allow us to *soften* the max operator. We choose to focus on soft versions of the three classical Pythagorean means parameterized by $\lambda$ where $\lambda = 0$ corresponds to the mean and the max is recovered as $\lambda \to \infty$:

$$\text{AM}_{soft}(V, \lambda) = \sum_i^N w_i V_i \tag{3}$$

$$\text{GM}_{soft}(V, \lambda) = -\exp\Big(\sum_i^N w_i \log(-V_i)\Big) \tag{4}$$

$$\text{HM}_{soft}(V, \lambda) = \Big(\sum_i^N w_i V_i^{-1}\Big)^{-1} \tag{5}$$

where $w_i = e^{\lambda V_i} / \Sigma_j e^{\lambda V_j}$ with $\lambda \geq 0, V_i < 0$. Using a *softmax* also has the well known advantage of being differentiable (as opposed to subdifferentiable for max). Note that we only require continuity to guarantee that computing the *softmax* is actually equivalent to computing $V(\tilde{D}, G)$ where $\tilde{D}$ is some convex combination of $D_i$ (see Appendix A.5).

## 4.2 Using the Original Minimax Objective

To illustrate the effect the *softmax* has on training, observe that the component of $AM_{soft}(V, 0)$ relevant to generator training can be rewritten as

$$\frac{1}{N} \sum_i^N \mathbb{E}_{x \sim p_G(x)}\Big[\log(1 - D_i(x))\Big] = \frac{1}{N} \mathbb{E}_{x \sim p_G(x)}\Big[\log(z)\Big]. \tag{6}$$

where $z = \prod_i^N (1 - D_i(x))$. Note that the generator gradient, $|\frac{\partial \log(z)}{\partial z}|$, is minimized at $z = 1$ over $z \in (0, 1]$[1]. From this form, it is clear that $z = 1$ if and only if $D_i = 0 \, \forall i$, so $G$ only receives a vanishing gradient if all $D_i$ agree that the sample is fake; this is especially unlikely for large $N$. In other words, $G$ only needs to fool a single $D_i$ to receive constructive feedback. This result allows the generator to successfully minimize the original generator objective, $\log(1 - D)$. This is in contrast to the more popular $-\log(D)$ introduced to artificially enhance gradients at the start of training.

At the beginning of training, when $\max_{D_i} V(D_i, G)$ is likely too harsh a critic for the generator, we can set $\lambda$ closer to zero to use the mean, increasing the odds of providing constructive feedback to the generator. In addition, the discriminators have the added benefit of functioning as an ensemble, reducing the variance of the feedback presented to the generator, which is especially important when the discriminators are far from optimal and are still learning a reasonable decision boundary. As training progresses and the discriminators improve, we can increase $\lambda$ to become more critical of the generator for more refined training.

## 4.3 Maintaining Multiple Hypotheses

We argue for this ensemble approach on a more fundamental level as well. Here, we draw on the density ratio estimation perspective of GANs (Uehara et al. (2016)). The original GAN proof assumes we have access to $p_{data}(x)$, if only implicitly. In most cases of interest, the discriminator only has access to a finite dataset sampled from $p_{data}(x)$; therefore, when computing expectations of $V(D, G)$, we only draw samples from our finite dataset. This is equivalent to training a GAN with $p_{data}(x) = \tilde{p}_{data}$ which is a distribution consisting of point masses on all the data points in the dataset. For the sake of argument, let's assume we are training a discriminator and generator, each with infinite capacity. In this case, the global optimum ($p_G(x) = \tilde{p}_{data(x)}$) fails to capture any of the interesting structure from $p_{data}(x)$, the true distribution we are trying to learn. Therefore, it is actually critical that we avoid this global optimum.

---

[1] $\nabla_G V = -\sum_i \frac{D_i}{z} \frac{\partial D_i}{\partial G} \prod_{j \neq i} (1 - D_j) = -\frac{1}{z} \frac{\partial D_k}{\partial G}$ for $D_k = 1, D_{\neq k} = 0$. Our argument ignores $\frac{\partial D_k}{\partial G}$.

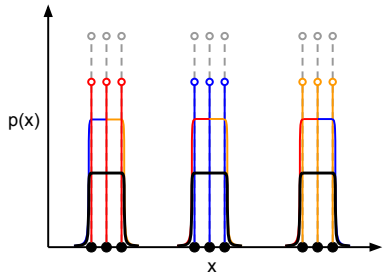

Figure 2: Consider a dataset consisting of the nine 1-dimensional samples in black. Their corresponding probability mass function is given in light gray. After training GMAN, three discriminators converge to distinct local optima which implicitly define distributions over the data (red, blue, yellow). Each discriminator may specialize in discriminating a region of the data space (placing more diffuse mass in other regions). Averaging over the three discriminators results in the distribution in black, which we expect has higher likelihood under reasonable assumptions on the structure of the true distribution.

In practice, this degenerate result is avoided by employing learners with limited capacity and corrupting data samples with noise (i.e., dropout), but we might better accomplish this by simultaneously training a variety of limited capacity discriminators. With this approach, we might obtain a diverse set of seemingly tenable hypotheses for the true $p_{data}(x)$. Averaging over these multiple locally optimal discriminators increases the entropy of $\tilde{p}_{data}(x)$ by diffusing the probability mass over the data space (see Figure 2 for an example).

### 4.4 Automating Regulation

The problem of keeping the discriminator and generator in balance has been widely recognized in previous work with GANs. Issues with unstable dynamics, oscillatory behavior, and generator collapse are not uncommon. In addition, the discriminator is often times able to achieve a high degree of classification accuracy (producing a single scalar) before the generator has made sufficient progress on the arguably more difficult generative task (producing a high dimensional sample). Salimans et al. (2016) suggested label smoothing to reduce the vulnerability of the generator to a relatively superior discriminator. Here, we explore an approach that enables the generator to automatically temper the performance of the discriminator when necessary, but still encourages the generator to challenge itself against more accurate adversaries. Specifically, we augment the generator objective:

$$\min_{G, \lambda > 0} F_G(V_i) - f(\lambda) \tag{7}$$

where $f(\lambda)$ is monotonically increasing in $\lambda$ which appears in the *softmax* equations, (3)—(5). In experiments, we simply set $f(\lambda) = c\lambda$ with $c$ a constant (e.g., 0.001). The generator is incentivized to increase $\lambda$ to reduce its objective at the expense of competing against the best available adversary $D^*$ (see Appendix A.6).

## 5 Evaluation

Evaluating GANs is still an open problem. In their original work, Goodfellow et al. (2014) report log likelihood estimates from Gaussian Parzen windows, which they admit, has high variance and is known not to perform well in high dimensions. Theis et al. (2016) recommend avoiding Parzen windows and argue that generative models should be evaluated with respect to their intended application. Salimans et al. (2016) suggest an *Inception score*, however, it assumes labels exist for the dataset. Recently, Im et al. (2016) introduced the Generative Adversarial Metric (GAM) for making pairwise comparisons between independently trained GAN models. The core idea behind their approach is given two generator, discriminator pairs ($G_1, D_1$) and ($G_2, D_2$), we should be able to learn their relative performance by judging each generator under the opponent's discriminator.

### 5.1 Metric

In GMAN, the opponent may have multiple discriminators, which makes it unclear how to perform the swaps needed for GAM. We introduce a variant of GAM, the generative multi-adversarial metric (GMAM), that is amenable to training with multiple discriminators,

$$\text{GMAM} = \log\Big(\frac{F^a_{G_b}(V^a_i)}{F^a_{G_a}(V^a_i)}\Big/\frac{F^a_{G_a}(V^b_i)}{F^b_{G_b}(V^b_i)}\Big).$$ (8)

where $a$ and $b$ refer to the two GMAN variants (see Section 3 for notation $F_G(V_i)$). The idea here is similar. If $G_2$ performs better than $G_1$ with respect to both $D_1$ and $D_2$, then GMAM>0 (remember $V \leq 0$ always). If $G_1$ performs better in both cases, GMAM<0, otherwise, the result is indeterminate.

### 5.2 Experiments

We evaluate the aforementioned variations of GMAN on a variety of image generation tasks: MNIST (LeCun et al. (1998)), CIFAR-10 (Krizhevsky (2009)) and CelebA (Liu et al. (2015)). We focus on rates of convergence to steady state along with quality of the steady state generator according to the GMAM metric. To summarize, loosely in order of increasing discriminator leniency, we compare

- F-boost: A single *AdaBoost.OL*-boosted discriminator (see Appendix A.7).
- P-boost: $D_i$ is trained according to *AdaBoost.OL*. A max over the weak learner losses is presented to the generator instead of the boosted prediction (see Appendix A.7).
- GMAN-max: $\max\{V_i\}$ is presented to the generator.
- GAN: Standard GAN with a single discriminator (see Appendix A.2).
- mod-GAN: GAN with modified objective (generator minimizes $-\log(D(G(z)))$).
- GMAN-$\lambda$: GMAN with $F :=$ arithmetic *softmax* with parameter $\lambda$.
- GMAN*: The arithmetic *softmax* is controlled by the generator through $\lambda$.

All generator and discriminator models are deep (de)convolutional networks (Radford et al. (2015)), and aside from the boosted variants, all are trained with Adam (Kingma & Ba (2014)) and batch normalization (Ioffe & Szegedy (2015)). Discriminators convert the real-valued outputs of their networks to probabilities with *squashed*-sigmoids to prevent saturating logarithms in the minimax objective ($\epsilon + \frac{1-2\epsilon}{1+e^{-x}}$). See Appendix A.8 for further details. We test GMAN systems with $N = \{2, 5\}$ discriminators. We maintain discriminator diversity by varying dropout and network depth.

#### 5.2.1 MNIST

Figure 3 reveals that increasing the number of discriminators reduces the number of iterations to steady-state by 2x on MNIST; increasing $N$ (the size of the discriminator ensemble) also has the added benefit of reducing the variance the minimax objective over runs. Figure 4 displays the variance of the same objective over a sliding time window, reaffirming GMAN's acceleration to steady-state. Figure 5 corroborates this conclusion with recognizable digits appearing approximately an epoch before the single discriminator run; digits at steady-state appear slightly sharper as well.

Our GMAM metric (see Table 1) agrees with the relative quality of images in Figure 5 with GMAN* achieving the best overall performance. Figure 6 reveals GMAN*'s attempt to regulate the difficulty of the game to accelerate learning. Figure 7 displays the GMAM scores comparing fixed $\lambda$'s to the variable $\lambda$ controlled by GMAN*.

| | Score | Variant | GMAN* | GMAN-0 | GMAN-max | mod-GAN |
|---|---|---|---|---|---|---|
| Better→ | **0.127** | GMAN* | - | $-0.020 \pm 0.009$ | $-0.028 \pm 0.019$ | $-0.089 \pm 0.036$ |
| | 0.007 | GMAN-0 | $0.020 \pm 0.009$ | - | $-0.013 \pm 0.015$ | $-0.018 \pm 0.027$ |
| | $-0.034$ | GMAN-max | $0.028 \pm 0.019$ | $0.013 \pm 0.015$ | - | $-0.011 \pm 0.024$ |
| | $-0.122$ | mod-GAN | $0.089 \pm 0.036$ | $0.018 \pm 0.027$ | $0.011 \pm 0.024$ | - |

Table 1: Pairwise GMAM metric means with *stdev* for select models on MNIST. For each column, a positive GMAM indicates better performance relative to the row opponent; negative implies worse. Scores are obtained by summing each variant's column.

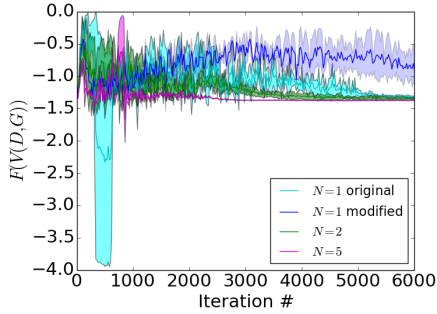
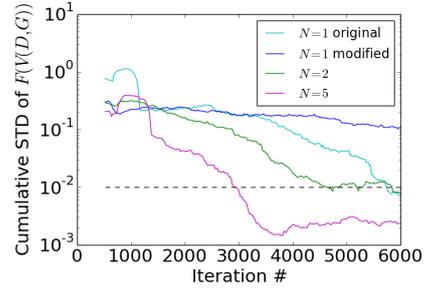

Figure 3: Generator objective, $F$, averaged over 5 training runs on MNIST. Increasing the number of discriminators accelerates convergence of $F$ to steady state (solid line) and reduces its variance, $\sigma^2$ (filled shadow $\pm 1\sigma$). Figure 4 provides alternative evidence of GMAN*'s accelerated convergence.

Figure 4: *Stdev*, $\sigma$, of the generator objective over a sliding window of 500 iterations. Lower values indicate a more steady-state. GMAN* with $N = 5$ achieves steady-state at $\approx 2\text{x}$ speed of GAN ($N = 1$). Note Figure 3's filled shadows reveal *stdev* of $F$ over runs, while this plot shows *stdev* over time.

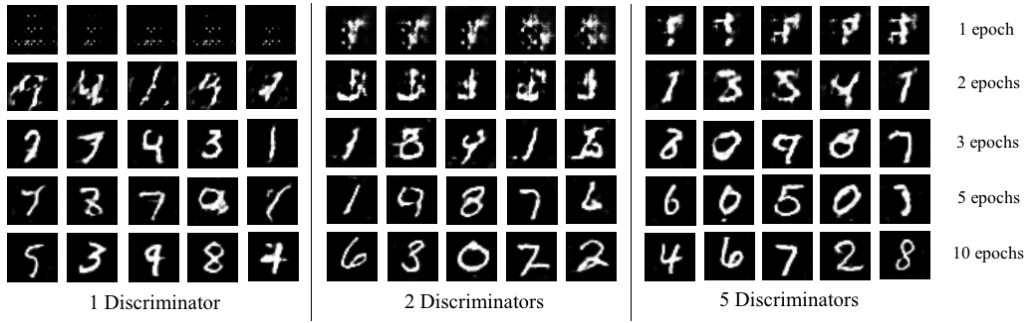

Figure 5: Comparison of image quality across epochs for $N = \{1, 2, 5\}$ using GMAN-0 on MNIST.

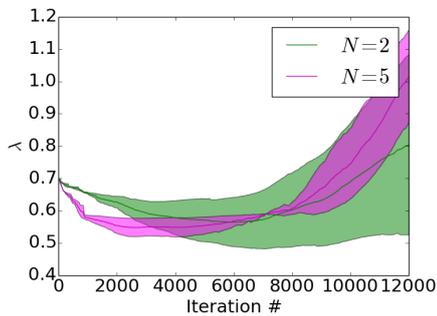

Figure 6: GMAN* regulates difficulty of the game by adjusting $\lambda$. Initially, $G$ reduces $\lambda$ to ease learning and then gradually increases $\lambda$ for a more challenging learning environment.

Figure 7: Pairwise $\frac{\text{GMAM}}{stdev(\text{GMAM})}$ for GMAN-$\lambda$ and GMAN* ($\lambda^*$) over 5 runs on MNIST.

| | Score | $\lambda$ ($N = 5$) | $\lambda^*$ | $\lambda = 1$ | $\lambda = 0$ |
|---|---|---|---|---|---|
| Better→ | **0.028** | $\lambda^*$ | - | $\frac{-0.008}{\pm 0.009}$ | $\frac{-0.019}{\pm 0.010}$ |
| | 0.001 | $\lambda = 1$ | $\frac{0.008}{\pm 0.009}$ | - | $\frac{-0.008}{\pm 0.010}$ |
| | −0.025 | $\lambda = 0$ | $\frac{0.019}{\pm 0.010}$ | $\frac{0.008}{\pm 0.010}$ | - |

### 5.2.2 CELEBA & CIFAR-10

We see similar accelerated convergence behavior for the CelebA dataset in Figure 8.

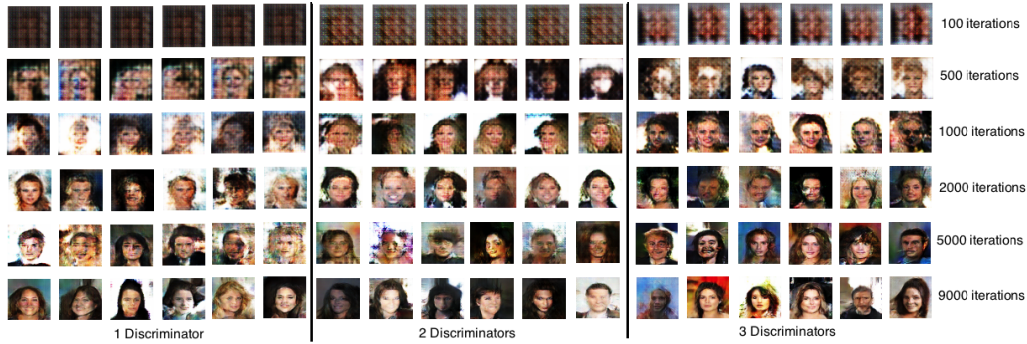

Figure 8: Image quality improvement across number of generators at same number of iterations for GMAN-0 on CelebA.

Figure 9 displays images generated by GMAN-0 on CIFAR-10. See Appendix A.3 for more results.

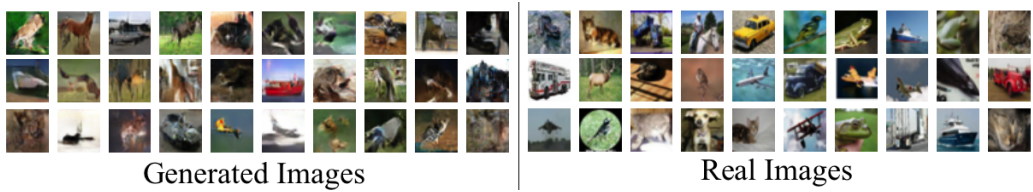

Figure 9: Images generated by GMAN-0 on the CIFAR-10 dataset.

We also found that GMAN is robust to *mode collapse*. We believe this is because the generator must appease a diverse set of discriminators in each minibatch. Emitting a single sample will score well for one discriminator at the expense of the rest of the discriminators. Current solutions (e.g., minibatch discrimination) are quadratic in batch size. GMAN, however, is linear in batch size.

## 6 CONCLUSION

We introduced multiple discriminators into the GAN framework and explored discriminator roles ranging from a formidable adversary to a forgiving teacher. Allowing the generator to automatically tune its learning schedule (GMAN*) outperformed GANs with a single discriminator on MNIST. In general, GMAN variants achieved faster convergence to a higher quality steady state on a variety of tasks as measured by a GAM-type metric (GMAM). In addition, GMAN makes using the original GAN objective possible by increasing the odds of the generator receiving constructive feedback.

In future work, we will look at more sophisticated mechanisms for letting the generator control the game as well as other ways to ensure diversity among the discriminators. Introducing multiple generators is conceptually an obvious next step, however, we expect difficulties to arise from more complex game dynamics. For this reason, game theory and game design will likely be important.

### ACKNOWLEDGMENTS

We acknowledge helpful conversations with Stefan Dernbach, Archan Ray, Luke Vilnis, Ben Turtel, Stephen Giguere, Rajarshi Das, and Subhransu Maji. We also thank NVIDIA for donating a K40 GPU. This material is based upon work supported by the National Science Foundation under Grant Nos. IIS-1564032. Any opinions, findings, and conclusions or recommendations expressed in this material are those of the authors and do not necessarily reflect the views of the NSF.

# A Appendix

## A.1 Accelerated Convergence & Reduced Variance

See Figures 10, 11, 12, and 13.

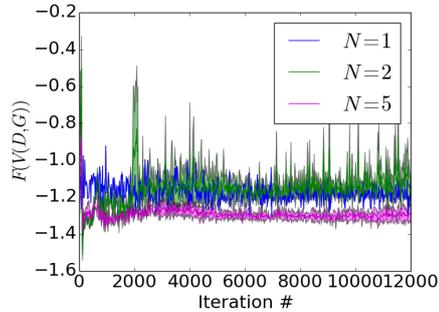

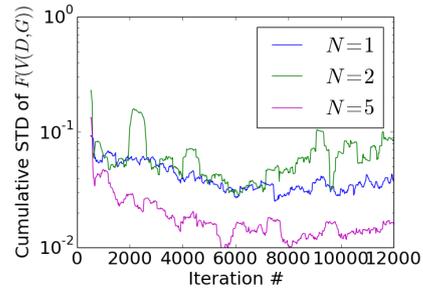

Figure 10: Generator objective, $F$, averaged over 5 training runs on CelebA. Increasing $N$ (# of $D$) accelerates convergence of $F$ to steady state (solid line) and reduces its variance, $\sigma^2$ (filled shadow $\pm 1\sigma$). Figure 11 provides alternative evidence of GMAN-0's accelerated convergence.

Figure 11: *Stdev*, $\sigma$, of the generator objective over a sliding window of 500 iterations. Lower values indicate a more steady-state. GMAN-0 with $N = 5$ achieves steady-state at $\approx$2x speed of GAN ($N = 1$). Note Figure 10's filled shadows reveal *stdev* of $F$ over runs, while this plot shows *stdev* over time.

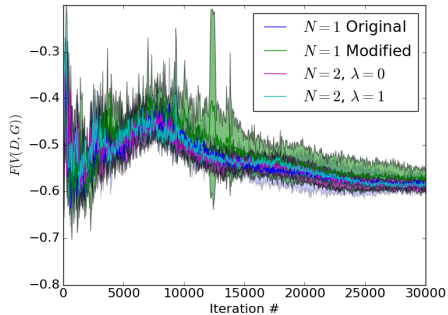

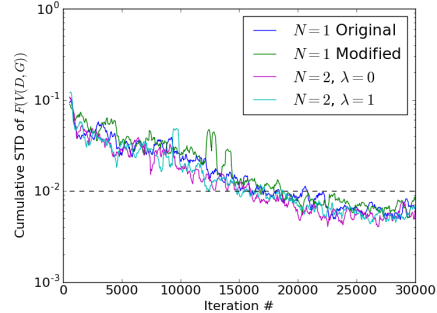

Figure 12: Generator objective, $F$, averaged over 5 training runs on CIFAR-10. Increasing $N$ (# of $D$) accelerates convergence of $F$ to steady state (solid line) and reduces its variance, $\sigma^2$ (filled shadow $\pm 1\sigma$). Figure 13 provides alternative evidence of GMAN-0's accelerated convergence.

Figure 13: *Stdev*, $\sigma$, of the generator objective over a sliding window of 500 iterations. Lower values indicate a more steady-state. GMAN-0 with $N = 5$ achieves steady-state at $\approx$2x speed of GAN ($N = 1$). Note Figure 12's filled shadows reveal *stdev* of $F$ over runs, while this plot shows *stdev* over time.

## A.2 Additional GMAM Tables

See Tables 2, 3, 4, 5, 6. Increasing the number of discriminators from 2 to 5 on CIFAR-10 significantly improves scores over the standard GAN both in terms of the GMAM metric and Inception scores.

## A.3 Generated Images

See Figures 14 and 15.

| | Score | Variant | GMAN* | GMAN-1 | GAN | GMAN-0 | GMAN-max | mod-GAN |
|---|---|---|---|---|---|---|---|---|
| Better→ | **0.184** | GMAN* | - | −0.007 | −0.040 | −0.020 | −0.028 | −0.089 |
| | 0.067 | GMAN-1 | 0.007 | - | −0.008 | −0.008 | −0.021 | −0.037 |
| | 0.030 | GAN | 0.040 | 0.008 | - | 0.002 | −0.018 | −0.058 |
| | 0.005 | GMAN-0 | 0.020 | 0.008 | 0.002 | - | −0.013 | −0.018 |
| | −0.091 | GMAN-max | 0.028 | 0.021 | 0.018 | 0.013 | - | −0.011 |
| | −0.213 | mod-GAN | 0.089 | 0.037 | 0.058 | 0.018 | 0.011 | - |

Table 2: Pairwise GMAM metric means for select models on MNIST. For each column, a positive GMAM indicates better performance relative to the row opponent; negative implies worse. Scores are obtained by summing each column.

| | Score | Variant | GMAN-0 | GMAN-1 | GMAN* | mod-GAN |
|---|---|---|---|---|---|---|
| Better→ | **0.172** | GMAN-0 | - | −0.022 | −0.062 | −0.088 |
| | 0.050 | GMAN-1 | 0.022 | - | 0.006 | −0.078 |
| | −0.055 | GMAN* | 0.062 | −0.006 | - | −0.001 |
| | −0.167 | mod-GAN | 0.088 | 0.078 | 0.001 | - |

Table 3: Pairwise GMAM metric means for select models on CIFAR-10. For each column, a positive GMAM indicates better performance relative to the row opponent; negative implies worse. Scores are obtained by summing each column. GMAN variants were trained with **two** discriminators.

| | GMAN-0 | GMAN-1 | mod-GAN | GMAN* |
|---|---|---|---|---|
| Score | **5.878** ± 0.193 | 5.765 ± 0.168 | 5.738 ± 0.176 | 5.539 ± 0.099 |

Table 4: Inception score means with standard deviations for select models on CIFAR-10. Higher scores are better. GMAN variants were trained with **two** discriminators.

| | Score | Variant | GMAN-0 | GMAN* | GMAN-1 | mod-GAN |
|---|---|---|---|---|---|---|
| Better→ | **0.180** | GMAN-0 | - | −0.008 | −0.041 | −0.132 |
| | 0.122 | GMAN* | 0.008 | - | −0.038 | −0.092 |
| | 0.010 | GMAN-1 | 0.041 | 0.038 | - | −0.089 |
| | −0.313 | mod-GAN | 0.132 | 0.092 | 0.089 | - |

Table 5: Pairwise GMAM metric means for select models on CIFAR-10. For each column, a positive GMAM indicates better performance relative to the row opponent; negative implies worse. Scores are obtained by summing each column. GMAN variants were trained with **five** discriminators.

| | GMAN-1 | GMAN-0 | GMAN* | mod-GAN |
|---|---|---|---|---|
| Score | **6.001** ± 0.194 | 5.957 ± 0.135 | 5.955 ± 0.153 | 5.738 ± 0.176 |

Table 6: Inception score means with standard deviations for select models on CIFAR-10. Higher scores are better. GMAN variants were trained with **five** discriminators.

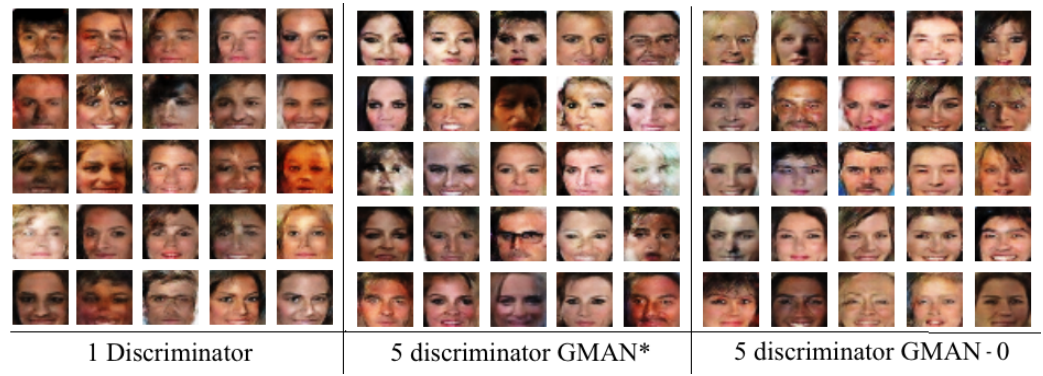

1 Discriminator    |    5 discriminator GMAN*    |    5 discriminator GMAN·0

Figure 14: Sample of pictures generated on CelebA cropped dataset.

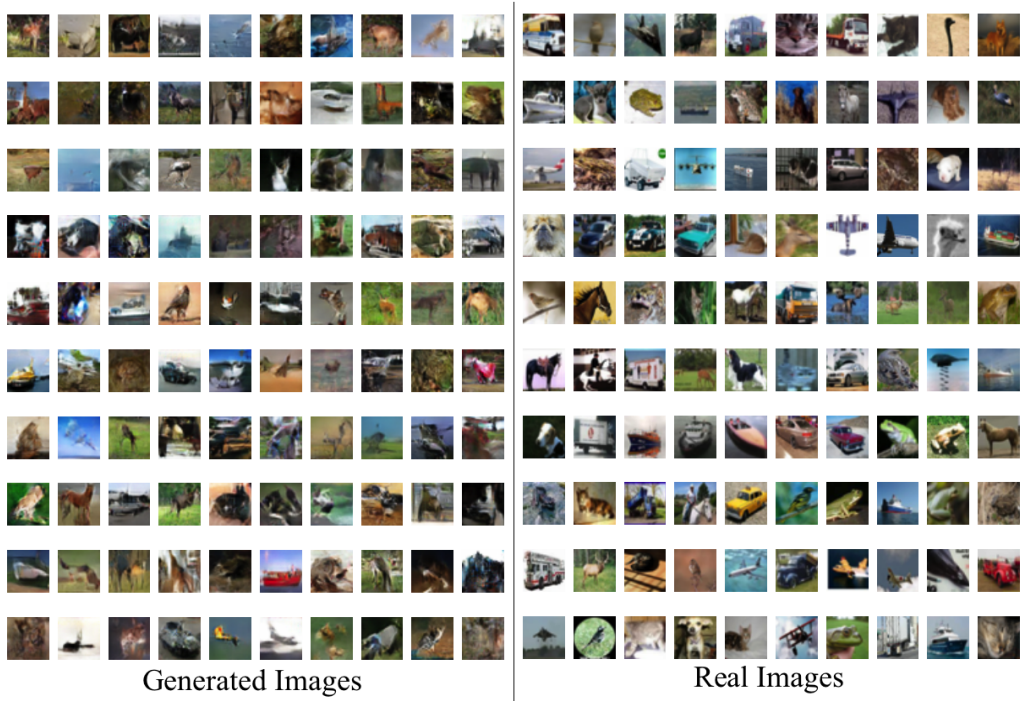

Generated Images        Real Images

Figure 15: Sample of pictures generated by GMAN-0 on CIFAR dataset.

### A.4 SOMEWHAT RELATED WORK

A GAN framework with two discriminators appeared in Yoo et al. (2016), however, it is applicable only in a semi-supervised case where a label can be assigned to subsets of the dataset (e.g., $\mathcal{X} = \{\mathcal{X}_1 = \text{Domain 1}, \mathcal{X}_2 = \text{Domain 2}, \ldots\}$). In contrast, our framework applies to an unsupervised scenario where an obvious partition of the dataset is unknown. Furthermore, extending GMAN to the semi-supervised domain-adaptation scenario would suggest multiple discriminators per domain, therefore our line of research is strictly orthogonal to that of their multi-domain discriminator approach. Also, note that assigning a discriminator to each domain is akin to prescribing a new discriminator to each value of a conditional variable in conditional GANs (Mirza & Osindero (2014)). In this case, we interpret GMAN as introducing multiple conditional discriminators and not a discriminator for each of the possibly exponentially many conditional labels.

In Section 4.4, we describe an approach to customize adversarial training to better suit the development of the generator. An approach with similar conceptual underpinnings was described in Ravanbakhsh et al. (2016), however, similar to the above, it is only admissible in a semi-supervised scenario whereas our applies to the unsupervised case.

### A.5 *Softmax* REPRESENTABILITY

Let $softmax(V_i) = \hat{V} \in [\min_{V_i}, \max_{V_i}]$. Also let $a = arg\min_i V_i$, $b = arg\max_i V_i$, and $\mathcal{V}(t) = V((1-t)D_a + tD_b)$ so that $\mathcal{V}(0) = V_a$ and $\mathcal{V}(1) = V_b$. The *softmax* and minimax objective $V(D_i, G)$ are both continuous in their inputs, so by the *intermediate value theorem*, we have that $\exists \, \hat{t} \in [0, 1] \; s.t. \; \mathcal{V}(\hat{t}) = \hat{V}$, which implies $\exists \, \hat{D} \in \mathcal{D} \; s.t. \; V(\hat{D}, G) = \hat{V}$. This result implies that the *softmax* (and any other continuous substitute) can be interpreted as returning $V(\hat{D}, G)$ for some $\hat{D}$ selected by computing an another, unknown function over the space of the discriminators. This result holds even if $\hat{D}$ is not representable by the architecture chosen for $D$'s neural network.

### A.6 Unconstrained Optimization

To convert GMAN* minimax formulation to an unconstrained minimax formulation, we introduce an auxiliary variable, $\Lambda$, define $\lambda(\Lambda) = \log(1 + e^\Lambda)$, and let the generator minimize over $\Lambda \in \mathbb{R}$.

### A.7 Boosting with *AdaBoost.OL*

*AdaBoost.OL* (Beygelzimer et al. (2015)) does not require knowledge of the weak learner's slight edge over random guessing ($P(\text{correct label}) = 0.5 + \gamma \in (0, 0.5]$), and in fact, allows $\gamma < 0$. This is crucial because our weak learners are deep nets with unknown, possibly negative, $\gamma$'s.

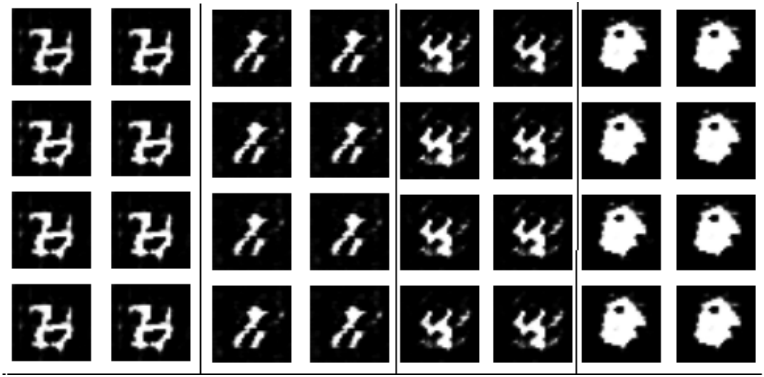

Figure 16: Sample of pictures generated across 4 independent runs on MNIST with F-boost (similar results with P-boost).

### A.8 Experimental Setup

All experiments were conducted using an architecture similar to DCGAN (Radford et al. (2015)). We use convolutional transpose layers (Zeiler et al. (2010)) for $G$ and strided convolutions for $D$ except for the input of $G$ and the last layer of $D$. We use the single step gradient method as in (Nowozin et al. (2016)), and batch normalization (Ioffe & Szegedy (2015)) was used in each of the generator layers. The different discriminators were trained with varying dropout rates from $[0.3, 0.7]$. Variations in the discriminators were effected in two ways. We varied the architecture by varying the number of filters in the discriminator layers (reduced by factors of 2, 4 and so on), as well as varying dropout rates. Secondly we also decorrelated the samples that the discriminators were training on by splitting the minibatch across the discriminators. The code was written in Tensorflow (Abadi et al. (2016)) and run on Nvidia GTX 980 GPUs. Code to reproduce experiments and plots is at https://github.com/iDurugkar/GMAN. Specifics for the MNIST architecture and training are:

- Generator latent variables $z \sim \mathcal{U}(-1, 1)^{100}$
- Generator convolution transpose layers: $(4, 4, 128), (8, 8, 64), (16, 16, 32), (32, 32, 1)$
- Base Discriminator architecture: $(32, 32, 1), (16, 16, 32), (8, 8, 64), (4, 4, 128)$.
- Variants have either convolution 3 $(4, 4, 128)$ removed or all the filter sizes are divided by 2 or 4. That is, $(32, 32, 1), (16, 16, 16), (8, 8, 32), (4, 4, 64)$ or $(32, 32, 1), (16, 16, 8), (8, 8, 16), (4, 4, 32)$.
- ReLu activations for all the hidden units. Tanh activation at the output units of the generator. Sigmoid at the output of the Discriminator.
- Training was performed with Adam (Kingma & Ba (2014)) ($lr = 2 \times 10^{-4}$, $\beta_1 = 0.5$).
- MNIST was trained for 20 epochs with a minibatch of size 100.
- CelebA and CIFAR were trained over 24000 iterations with a minibatch of size 100.

14